\documentclass[runningheads]{llncs}
\usepackage{graphicx}

\usepackage{tikz}
\usepackage{comment}
\usepackage{amsmath,amssymb} 
\usepackage{color}
\usepackage{booktabs}
\usepackage{multirow}
\usepackage{dsfont}
\usepackage{wrapfig}
\usepackage[pagebackref,breaklinks,colorlinks]{hyperref}
\usepackage{orcidlink}
\usepackage[accsupp]{axessibility}  

\begin{document}
\pagestyle{headings}
\mainmatter
\def\ECCVSubNumber{7016}  

\title{Single-Stream Multi-Level Alignment for Vision-Language Pretraining} 


\titlerunning{}
%

\author{Zaid Khan\orcidlink{0000-0003-0743-2992} \inst{1} \and
Vijay Kumar B G\orcidlink{0000-0001-8188-2241}\inst{2} \and
Xiang Yu\orcidlink{0000-0003-2765-2749} \inst{2} \and
Samuel Schulter\orcidlink{0000-0002-7341-5907} \inst{2} \and
Manmohan Chandraker\orcidlink{0000-0003-4683-2454} \inst{2,3}\and
Yun Fu\orcidlink{0000-0002-5098-2853} \inst{1}}
%
\authorrunning{Z. Khan et al.}
%
\institute{ Northeastern University \and  NEC Labs America \and  UC San Diego \\
\email{khan.za@northeastern.edu, vijay.kumar@nec-labs.com,  xiangyu@nec-labs.com, samuel@nec-labs.com, mkchandraker@eng.ucsd.edu, yunfu@ece.neu.edu}
}
\maketitle

\begin{abstract}
Self-supervised vision-language pretraining from pure images and text with a contrastive loss is effective, but ignores fine-grained alignment due to a dual-stream architecture that aligns image and text representations only on a global level.
Earlier, supervised, non-contrastive methods were capable of finer-grained alignment, but required dense annotations that were not scalable. 
We propose a single stream architecture that aligns images and language at multiple levels: global, fine-grained patch-token, and conceptual/semantic, using two novel tasks: symmetric cross-modality reconstruction (XMM) and a pseudo-labeled key word prediction (PSL). In XMM, we mask input tokens from one modality and use cross-modal information to reconstruct the masked token, thus improving fine-grained alignment between the two modalities. In PSL, we use attention to select keywords in a caption, use a momentum encoder to recommend other important keywords that are missing from the caption but represented in the image, and then train the visual encoder to predict the presence of those keywords, helping it learn semantic concepts that are essential for grounding a textual token to an image region. 
We demonstrate competitive performance and improved data efficiency on image-text retrieval, grounding, visual question answering/reasoning against larger models and models trained on more data. 
Code and models available at \url{zaidkhan.me/SIMLA}.

\end{abstract}

\section{Introduction}

To learn a join representation of images and language, early work \cite{uniter,ViLBERT,LXMERT} follows a supervised approach, using a pre-trained object detector to extract image regions, which are then aligned with corresponding image captions or dense annotations.  
Such approaches are limited by the amount of available densely annotated data and the semantic concepts the pretrained object detector can represent.
A recent alternative approach is to directly align image representations with the corresponding text representations using a contrastive loss \cite{clip,declip,align,triple_contrastive_learning,cyclip,slip}, sidestepping the need for a pretrained object detector or dense annotations.
Such approaches can learn from image-text pairs alone, which can be scraped from the web at large scales. 
However, the image-text contrastive learning paradigm is data hungry, using $1\mathrm{b}+$~\cite{align,coca} or $100\mathrm{m}+$~\cite{clip,lit} pairs to overcome the noisiness of web-scraped image-text pairs.
Second, the standard image-text contrastive learning architecture and objective uses a dual-stream architecture that aligns the global image and text representations, making it difficult to learn fine-grained details \cite{filip}.
Third, contrastive learning does not explicitly align visual and language \textit{concepts}, only features. 
Because the data complexity of images is greater than that of short captions, it can be challenging for the vision model to learn a representation that captures modality-invariant instance information corresponding to coherent natural language concepts rather than vision-specific semantics irrelevant to the modality-invariant image content. 

We propose an approach that aligns image and language representations on multiple levels using a single stream transformer-only architecture that enables early, local interactions between image regions and language tokens, without the need for a pretrained object detector or dense annotations. 
We design a a symmetric cross-modality reconstruction task to teach fine-grained alignment between image patches and language tokens, and construct a concept prediction task that extracts pseudo labels for each image without supervision and trains the visual encoder to detect concepts that are missing from the caption but present in the image. 
This allows us to align vision and language on multiple levels: fine-grained (cross-modality reconstruction), coarse (contrastive learning) and discrete (concept-level supervision). 
We empirically evaluate our proposed model, SIMLA (\textbf{SI}ngle-Stream \textbf{M}ulti-\textbf{L}evel \textbf{A}lignment) on several downstream tasks, following prior work~\cite{ALBEF}.
The entirely self-supervised SIMLA achieves state-of-the-art results on image-text retrieval and grounding, while outperforming larger models trained with supervision on downstream vision-language tasks, and demonstrates greater data efficiency compared to prior work in an ablation study.
Our contributions, summarized:
\vspace{-2mm}
\begin{enumerate}
    \item A symmetric cross-modality reconstruction task to learn fine-grained alignment between image patches and language tokens.
    \item A natural language, pseudo-labeling approach to align concept-level semantics without dense annotations.
    \item A single-stream architecture to enable the proposed multi-level alignment.
    \item Extensive experiments on image-text retrieval, vision-language reasoning, and visual grounding to demonstrate effectiveness of the proposed modules.
\end{enumerate}

\section{Method}
Images are dense, unstructured, and require significant processing to extract useful semantic information.
In contrast, language is highly structured, and contains directly available semantic information.
Because of this asymmetry, attempting to align image features with language features too early will be futile, because the image features are too low-level to be matched with the more abstract language features.
Contemporary architectures thus employ a symmetric encoder design, in which both image and text are processed by equally deep encoders before late fusion through alignment of global image and text representations.
This approach wastes model capacity, as high-level image semantics often correspond directly to low-level language semantics, so processing language to same depth as images is wasteful.
In addition, both language and images contain a semantic pyramid of concepts, with some concepts being highly localized (e.g. a small image patch / single word) while other concepts are highly abstract (e.g multiple interrelated image patches / multi-token sequences).
Cross-modal concepts can exist at different levels of the semantic pyramid for each modality (e.g the singular token 'throwing' describes a complex spatial scene / the phrase 'bird walking on rocky ground' may describe a small local image region). 
Thus, the problems in vision-language learning are twofold: 
\begin{enumerate}
    \item Asymmetry in inherent semantic abstraction between image and text data.
    \item Semantic concepts appear at disparate levels in the abstraction hierarchy across modalities.
\end{enumerate}

We propose an asymmetric architecture with a multi-task loss to address the above issues.
Concretely, our architecture consists of a deep stack of transformer encoder layers that can be interpreted as a transformer language model \cite{transformer} stacked atop a visual transformer \cite{vit}.
During the forward pass, an image is fed through the bottom of the stack, while language tokens are injected at the middle of the stack, into the bottom of the language model.
This design allows processing of the image to an appropriate level of semantic abstraction before fusion with language.
Our multi-task loss consists of four tasks, engineered to align vision and language representations at multiple levels.
We begin with an image-text matching task for very coarse instance-level alignment, and add a contrastive loss for global feature-level alignment.
Next, we add a patch-level reconstruction task for fine-grained region-level alignment.
Finally, we add a pseudo-label supervision task to the visual encoder to explicitly ensure the level of abstraction between the visual and language tokens is synchronized prior to fine-grained fusion.

\subsection{Preliminary Architectures}
\begin{figure}[t]
\includegraphics[width=\textwidth]{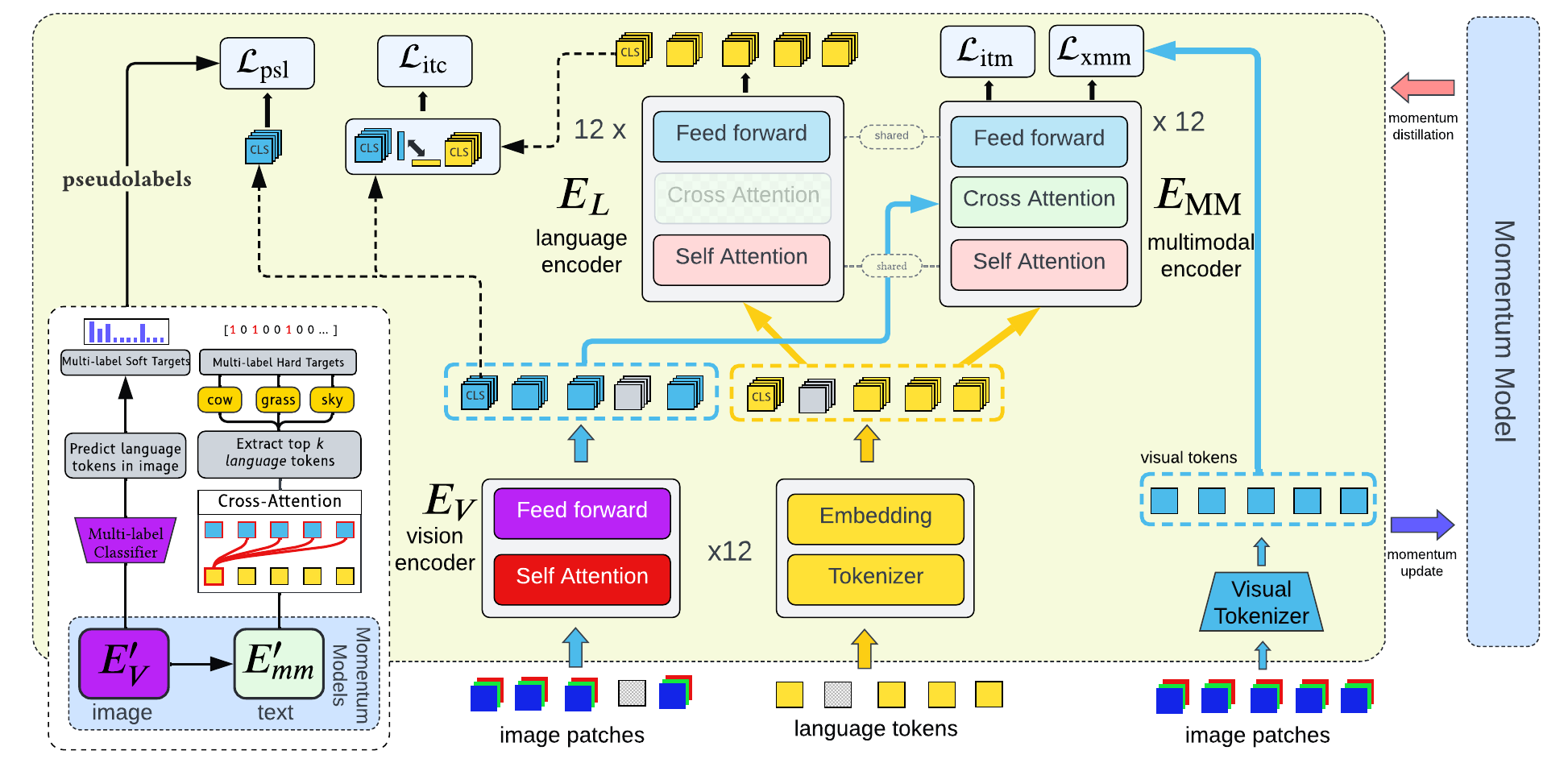}
\caption{
SIMLA architecture. 
A language encoder $E_{l}$ is stacked atop a vision encoder $E_v$.
We add cross attention to $E_{l}$, allowing us to reuse it as a multimodal encoder $E_{mm}$ by consuming image embeddings from $E_v$.
Four tasks align images and language at multiple levels, exploiting a momentum model for additional supervision.
A D-VAE tokenizes image patches for the cross-modality reconstruction task.
}
\end{figure}
Our model is a 24-deep stack of transformer \cite{transformer} layers that can be decomposed into a vision encoder $E_v$, a language encoder $E_l$, and a multimodal encoder $E_{mm}$.
Specifically, we stack the language encoder $E_l$ atop the vision encoder $E_v$.
We then add cross-attention layers after each self-attention layer in the language encoder $E_l$, allowing us to use it as a multimodal encoder $E_{mm}$ when an image-text pair is passed in, and as a unimodal language encoder when language tokens are passed in. 
To obtain a multimodal embedding, we first use the bottom half of the transformer encoder stack ($E_v$) to encode an input image $I$ into a sequence of embeddings
$E_v(I) = \{\Vec{v}_\mathrm{cls}, \Vec{v}_1,...,\Vec{v}_N\}$
where $v_\mathrm{cls}$ is the embedding of the \texttt{[CLS]} token.
We then pass the sequence of image embeddings $\{\Vec{v}_\mathrm{cls}, \Vec{v}_1,...,\Vec{v}_N\}$ into the top half of the transformer encoder stack, corresponding to the language model with cross-attention, while concurrently injecting the associated caption, so the image embeddings $\{\Vec{v}_\mathrm{cls}, \Vec{v}_1,...,\Vec{v}_N\}$ from bottom half of the stack and the input \textit{tokens} $\{\texttt{[cls]}, t_1,...,t_N\}$ are consumed simultaneously and fused through cross-attention after each self attention layer to yield a sequence of multimodal embeddings $\{\Vec{m}_\mathrm{cls}, \Vec{m}_1,...,\Vec{m}_N\} = E_{mm}(\{\Vec{v}_\mathrm{cls}, \Vec{v}_1,...,\Vec{v}_N\}, \{\texttt{[cls]}, t_1,...,t_N\})$.

\subsection{Coarse Cross-Modality Alignment}
\subsubsection{Image-Text Contrastive Learning}
The simplest level of alignment is coarse, global alignment between image and text representations.
Global alignment is useful training signal for two reasons: (i) it is robust to mismatches in fine-grained details between an image and caption (ii) it is an easier task than fine-grained alignment and enables faster learning during the earlier stages of training, when fine-grained alignment is infeasible due to the large domain gap between images and text. 
Coarse, global alignment requires learning image and text representations which capture modality-invariant information. 
A simple, effective and scalable \cite{clip,align} approach to learning modality invariant representations is multi-view contrastive learning \cite{contrastive_multiview_coding}. 
The multi-view contrastive objective pushes embeddings of matched image-text pairs together while pulling those of unmatched image-text pairs apart. 
Our contrastive loss follows the InfoNCE \cite{infoNCE} formulation.
Contrastive losses benefit from larger batch sizes, but batch sizes are bounded by GPU memory.
To increase effective batch size, we follow MoCo \cite{moco} by using memory queues of size $M$ for the unimodal image ($Q^\mathrm{img}$) and text ($Q^\mathrm{txt}$) features, as well as maintaining momentum (time-averaged) versions of the text and image encoders.
The normalized image-to-text and text-to-image similarity are calculated as
\begin{align}
\begin{split}
    \label{eqn:sim}
p_m^\mathrm{i2t}(I, Q^\mathrm{txt}) &= \frac{\exp (S(I,Q^\mathrm{txt}_m) / \tau)}{\sum_{m=1}^M \exp (S(I,Q^\mathrm{txt}_m)/ \tau)} \\
p_m^\mathrm{t2i}(T, Q^{\mathrm{img}}) &= \frac{\exp (S(T,Q^{\mathrm{img}}_m)/ \tau)}{\sum_{m=1}^M \exp (S(T,Q^{\mathrm{img}}_m)/ \tau)}
\end{split}
\end{align}
where $\tau$ is a learnable temperature parameter, $S(I,T)=g_v(\Vec{v}_\mathrm{cls}) g'_l(\Vec{l}'_\mathrm{cls})$
and $S(T,I) = g_l(\Vec{l}_\mathrm{cls})^T g'_v(\Vec{v}'_\mathrm{cls})$ are raw similarity scores between image and text \texttt{{[CLS]}} tokens, obtained by $E_v(I)$ and $E_l(T)$ respectively. The functions $g_v$ and $g_l$ are linear transformations that project the unimodal \texttt{[CLS]} embeddings of the image and text, respectively, to lower-dimensional representations, followed by normalization to unit length.
We use $g'_v(\Vec{v}'_\mathrm{cls})$ and $g'_l(\Vec{l}'_\mathrm{cls})$ to denote the momentum features, retrieved from the memory queues.
The boolean one-hot vectors $\Vec{y}^\mathrm{i2t}(I)$ and $\Vec{y}^\mathrm{t2i}(T)$ represent the ground-truth similarity, with the positive pair indicated by a $1$ and a $0$ for all negatives.
Then, the image-text contrastive loss is defined as the cross-entropy $\mathrm{H}$ between $\Vec{p}$ and $\Vec{y}$:
\begin{equation}
\label{eqn:itc}
\mathcal{L}_\mathrm{itc} = \frac{1}{2} \mathbb{E}_{(I,T)\sim D} \big[ \mathrm{H}(\Vec{y}^\mathrm{i2t}(I),\Vec{p}^\mathrm{i2t}(I)) + \mathrm{H}(\Vec{y}^\mathrm{t2i}(T),\Vec{p}^\mathrm{t2i}(T)) \big]
\end{equation}
The one-hot labels $\Vec{y}^\mathrm{i2t}(I)$ and $\Vec{y}^\mathrm{t2i}(T)$ penalize all predictions which do not match each image to the text it came paired with, and vice versa. 
However, one caption can potentially describe many different images, and similarly, many captions may match an image.
To avoid this noisy penalization, we soften the hard targets with soft targets generated by the momentum model, corresponding to knowledge distillation \cite{kd_hinton} with the momentum model as a teacher.
The complete loss can then be written as 
\begin{align}
       \mathcal{L}_{\text {itc }}^{\text {mod }} &= (1-\alpha) \mathcal{L}_{\text {itc }} + \alpha \mathcal{L}^\prime_{\mathrm{itc}} \\
    \mathcal{L}^\prime_{\mathrm{itc}} &= \frac{1}{2} \mathbb{E}_{(I, T) \sim D}\left[\mathrm{H}\left(p_{m}^{i 2 \mathrm{t}}(I), p^{i 2 \mathrm{t}}(I)\right)+\mathrm{H}\left({p}_{m}^{t 2 \mathrm{i}}(T), p^{t 2 \mathrm{i}}(T)\right)\right]
\end{align}
 where $p_{m}^{i 2 \mathrm{t}}(I)$ and ${p}_{m}^{t 2 \mathrm{i}}(T)$ is Equation \ref{eqn:sim} using only the momentum encoders.

\subsubsection{Image-Text Matching} 
is a binary classification task to predict if an image-text pair is matched. 
We define the ITM loss to be 
\begin{equation}
\label{eqn:itm}
\mathcal{L}_{\mathrm{itm}}=\mathbb{E}_{(I, T) \sim D} \mathrm{H}\left(\boldsymbol{y}^{\mathrm{itm}}, \boldsymbol{p}^{\mathrm{itm}}(I, T)\right)
\end{equation}
where $\boldsymbol{y}^{\mathrm{itm}}$ is a one-hot vector indicating whether the pair is matched or not, and $p^{\text {itm }}$ is a two-class probability vector predicted by a single fully connected layer on top of the multimodal \texttt{[CLS]} token.
We mine in-batch hard negatives for each image and text in a pair following ALBEF \cite{ALBEF}.

\subsection{Finer-Grained Cross-Modality Alignment} 
A contrastive loss such as $\mathcal{L}_{\mathrm{itc}}$ aligns the global image and text representations.
However, solely aligning the global representations while simultaneously fusing the image and text at the last possible opportunity makes it difficult to learn fine-grained correspondences, such as those between subregions of an image and subsequences of a caption.
We design a reconstruction task to teach a model fine-grained alignment between images and patches.
We mask the image, and force the model to reconstruct the masked image region from the remaining portion of the image using the caption as context.
We then reverse the reconstruction task, forcing the model to reconstruct masked language tokens from the remaining portion of the caption using the image as context.

Concretely, $(I, T)$ be an image text pair.
Let $\mathcal{M}_I$ be a mask for the image, generated following the masking strategy of BEiT \cite{beit}, and let $\mathcal{M}_T$ be the mask for the language tokens, generated following the masking strategy of BERT \cite{bert}.
We then generate\footnote{We depict the masking as a boolean operation for notational simplicity. The implementation follows the strategy of BEiT\cite{beit} and BERT\cite{bert} for $I$, $T$ respectively.} a masked image as $\hat{I} = I \odot \mathcal{M}_I$ and masked text as $\hat{T} =  T \odot \mathcal{M}_T$.
Then, the loss to be minimized is 
\begin{equation}
\mathcal{L}_{\mathrm{xmm}}=\mathbb{E}_{(I, \hat{T}) \sim D} \mathrm{H}\left(\boldsymbol{y}^{\mathrm{MLM}}, \boldsymbol{p}^{\mathrm{MLM}}(I, \hat{T})\right) + \mathbb{E}_{(\hat{I}, T) \sim D} \mathrm{H}\left(\boldsymbol{y}^{\mathrm{MIM}}, \boldsymbol{p}^{\mathrm{MIM}}(\hat{I}, T)\right)
\end{equation}
The cross-modality masked language modeling loss $\mathcal{L}_{\mathrm{xmm}}$ is a sum of two cross-entropy losses, where $\boldsymbol{y}^{\mathrm{MLM}}$ and $\boldsymbol{y}^{\mathrm{MIM}}$ indicate the ground-truth value of the masked language token and masked image token respectively, and $\boldsymbol{p}^{\mathrm{MLM}}(I, \hat{T})$, $\boldsymbol{p}^{\mathrm{MIM}}(I, \hat{T})$ represents the model's probability estimates of the masked language and image tokens respectively.
Because images are continuous, use the strategy of \cite{beit} to discretize the images into a sequence of tokens and mask them. 
We divide each image into patches and tokenize each patch with a discrete VAE \cite{d_vae} that maps each patch to one of 8192 visual tokens from a learned codebook.

In many cases, the ground-truth visual or language token can be plausibly replaced with an alternative. 
However, the ground truth target vectors are one-hot encoded and penalize any predictions that do not exactly match for the ground truth, even if they are plausible.
Furthermore, the image masking and language masking are random, so it is possible for non-content tokens (e.g. \textit{the, it}) or tokens that cannot be predicted well based on context to be masked.
To allow the model to learn even when the ground-truth target for the masked token cannot be reasonably predicted from context, we again use the momentum distillation strategy.
Specifically, we decompose $\mathcal{L}_\mathrm{xmm}$ into
\begin{equation}
    \mathcal{L}^{\mathrm{mod}}_{\mathrm{xmm}} = (1-\alpha) \mathcal{L}_{\text {MIM }} + \alpha \mathcal{L}^\prime_{\mathrm{MIM}} + (1-\alpha) \mathcal{L}_{\text {MLM }} + \alpha \mathcal{L}^\prime_{\mathrm{MLM}}
\end{equation}
where $\mathcal{L}^\prime_{\mathrm{MIM}} = \mathrm{H}\left(\boldsymbol{p}_{m}^{\mathrm{MIM}}, \boldsymbol{p}^{\mathrm{MIM}}(I, \hat{T})\right)$, $\mathcal{L}^\prime_{\mathrm{MLM}} = \mathrm{H}\left(\boldsymbol{p}_{m}^{\mathrm{MLM}}, \boldsymbol{p}^{\mathrm{MLM}}(I, \hat{T})\right)$ and $\boldsymbol{p}_{m}^{\mathrm{MLM}}$, $\boldsymbol{p}_{m}^{\mathrm{MLM}}$ are the softmax-normalized outputs of the MIM and MLM momentum prediction heads over the visual and language token distributions, respectively. 

\subsection{Concept-Level Alignment} 
Semantic concepts may appear at disparate levels in the abstraction hierarchy
across modalities.
A concept may be highly complex in the visual modality, while being expressible with a single token in the language modality, and vice versa.
This results in a concept-level mismatch between images and text.
Although an asymmetric architecture that subjects image inputs to greater processing than text inputs prior to fusion addresses the intrinsic disparity in the semantic abstraction between image and text data, it does not guarantee that the visual embeddings $E_{v}(I)=\left\{\vec{v}_{\mathrm{cls}}, \vec{v}_{1}, \ldots, \vec{v}_{N}\right\}$ express concepts that are commonly described with language, or even possible to describe with language.
Furthermore, it is possible that during the alignment process, the \textit{unimodal} representations may degrade, because the emphasis is only on alignment. 

To address this, we design a  high-level alignment task in which the visual representation is aligned to represent concepts expressible by the language encoder by teaching it to label images with language concepts associated to the image, which also maintains the quality of the unimodal visual representation.
We use the self-attention map of the multimodal \texttt{[CLS]} token to determine which language tokens within the text are most salient to the image-text pair.
We choose $k$ of the most salient tokens as pseudo-labels for the image, and generate a "hard" 2-D binary target vector $\mathbf{y}^{\mathrm{PSL}} \in \mathbb{R}^{V}$, where $V$ is the number of tokens known to the language model, and a 1 in the $[0][i]$-th position indicates the $i$-th token is a target pseudo-label and a 1 in the $[1][j]$-th position indicates the $j$-th token is not a target.
We seek to minimize  
\begin{equation}
\mathcal{L}_\mathrm{PSL}=-\frac{1}{V} \sum_{i=1}^{V} \mathbf{y}^{\mathrm{PSL}}_{i} \cdot \log \left(\sigma (\mathbf{p}^{\mathrm{PSL}}_{i})\right)+\left(1-\mathbf{y}^{\mathrm{PSL}}_{i}\right) \cdot \log \left(1-\sigma (\mathbf{p}^{\mathrm{PSL}}_{i})\right)
\end{equation}
where $\mathbf{p}^\mathrm{PSL}$ is the output of a single fully-connected layer placed atop the unimodal image $\texttt{[CLS]}$ token, $\sigma(\cdot)$ is a sigmoid function used the clamp the output of the fully-connected layer between 0 and 1, and $V$ is the number of tokens in the vocabulary of the tokenizer.
This corresponds to multi-label loss where the model is trained to predict which language concepts (corresponding to tokens) are present in the image, using only the image context. 
However, the binary pseudolabels $\mathbf{y}^{\mathrm{PSL}}$ may fail to capture relevant concepts in the image, because the caption typically only describes a small number of aspects of an image.
To provide a stronger self-supervisory signal, we use the momentum model as a teacher and minimize the K-L divergence between the predicted pseudolabels and the momentum pseudolabels.
This can be expressed as a distillation loss where $\mathbf{p}^{\prime\mathrm{PSL}}$ is the vector of momentum pseudolabel predictions.
\begin{equation}
\mathcal{L}^{\mathrm{mod}}_\mathrm{PSL}= (1 - \alpha) \mathcal{L}_{\mathrm{PSL}} -\frac{\alpha}{V} \sum_{i=1}^{V} \mathbf{p}^{\prime\mathrm{PSL}}_{i} \cdot \log \left(\sigma (\mathbf{p}^{\mathrm{PSL}}_{i})\right)+\left(1-\mathbf{p}^{\prime \mathrm{PSL}}_{i}\right) \cdot \log \left(1-\sigma (\mathbf{p}^{\mathrm{PSL}}_{i})\right)
\end{equation}

The full pre-training objective can be expressed as 
\begin{equation}
    \mathcal{L} =  \mathcal{L}_{\text {itc }}^{\text {mod }} + \mathcal{L}^{\mathrm{mod}}_{\mathrm{xmm}} + \mathcal{L}_{\mathrm{itm}} + \mathcal{L}^{\mathrm{mod}}_\mathrm{PSL}
\end{equation}

\subsection{Implementation Details}
We initialize the bottom 12 layers of the transformer encoder stack dedicated to vision (corresponding to the visual encoder) with the weights and architecture ViT/B-16 vision transformer \cite{vit}, which is equipped with self-attention only. 
We initialize the top 12 multimodal layers of the transformer encoder stack (corresponding to the shared text / multimodal encoder) with the weights and architecture of BERT \cite{bert}, with cross-attention.
We pre-train the model for 30 epochs on 8 NVIDIA A100 GPUs with a batch size of 512.
During pre-training, random $256 \times 256$ crops of images are used and input, and RandAugment \cite{randaug} is applied to the images with color changes removed, following \cite{ALBEF}.
We set the momentum coefficient to 0.995, and linearly scale the distillation coefficient $\alpha$ from $0 \rightarrow 0.4$ in the first epoch.
We use an $M=65,536$ length memory queue.
The AdamW \cite{adamw} optimizer is used to train the model, with a  weight decay of 0.02 and a cosine learning rate scheduler with a linear warmup to $1e^{-4}$ followed by a decay to $1e^{-5}$ in the subsequent epochs.

\section{Experiments}
\begin{figure}
    \centering
    \includegraphics[width=\textwidth]{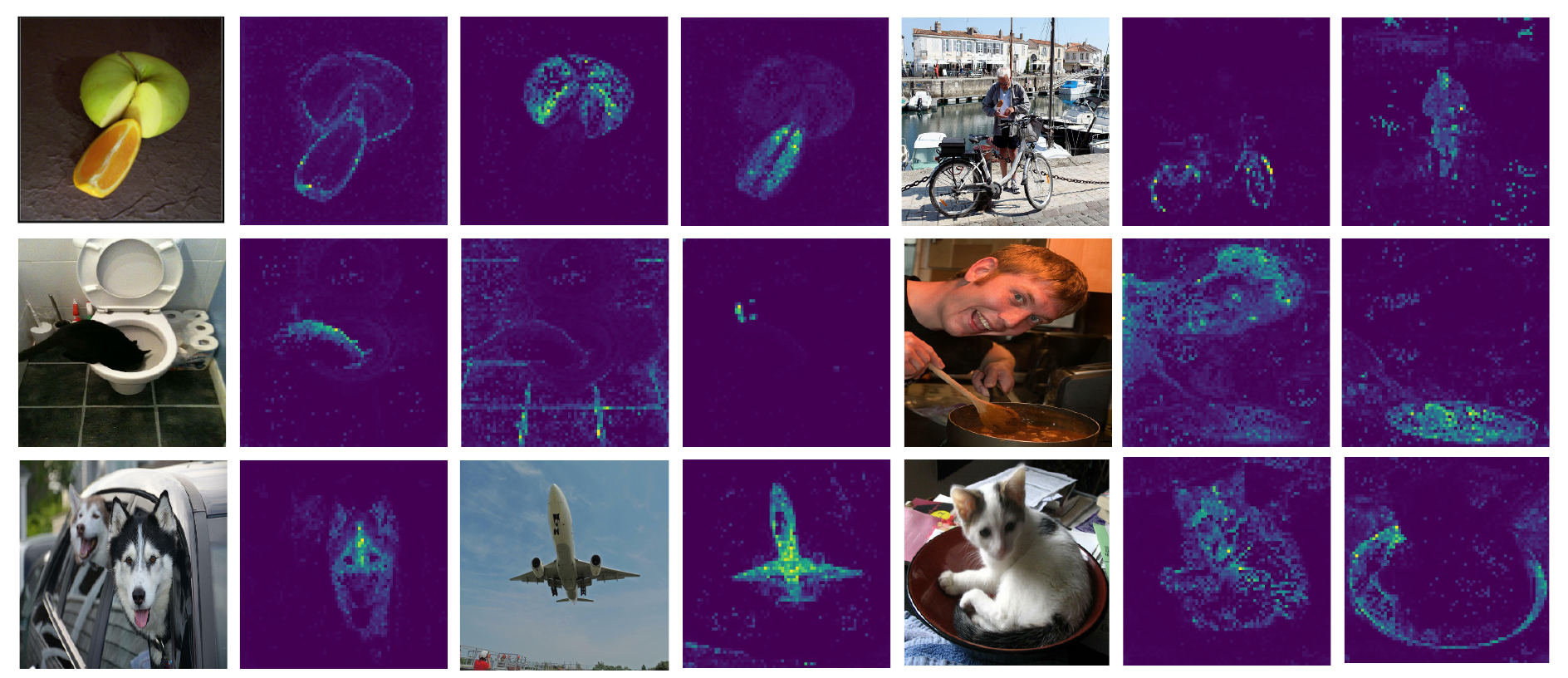}
    \caption{Self-attention maps of the visual \texttt{[CLS]} token from different heads.}
    \label{fig:vit-attention}
\end{figure}
\subsection{Experimental Setup}
\textbf{Pretraining Data}
is constructed by concatenating four image-text datasets: Conceptual Captions \cite{CC}, SBU Captions \cite{sbu}, COCO\cite{coco} and Visual Genome \cite{VG}, for a total of 4M image-text pairs, identical to \cite{uniter,ALBEF}.

\noindent\textbf{Image-Text Retrieval} The goal of text retrieval (TR) is to retrieve texts matching a query image.
Image retrieval (IR) reverses the roles of the modalities.
We evaluate retrieval on MSCOCO \cite{coco} and Flickr30k \cite{flickr}. 
We use the Karpathy\cite{karpathy} train/val/test splits for finetuning: 113k/5k/5k for MSCOCO and 29k/1k/1k for Flickr.
For $0$-shot retrieval on Flickr30k, we use the model fine-tuned on COCO, following \cite{ALBEF}.
We use ITC (Eq. \ref{eqn:itc}) and ITM losses (Eq. \ref{eqn:itm}) during fine-tuning.
We finetune using a learning rate of $1e-5$ for 10 epochs. 
For fashion image retrieval, we use FashionGen\cite{fashiongen}, following the protocol of \cite{fashionbert}.
Fashion retrieval results and evaluation details are in the appendix. 

\noindent\textbf{Visual Question Answering (VQA)} requires predicting an answer from an \texttt{(image, question)} pair. Following \cite{ALBEF}, we treat the task as an text generation problem using a auto-regressive decoder atop the multimodal encoder. For answer generation, we use \texttt{[CLS]} as the start of sequence token and \texttt{[SEP]} as the end of sequence token. 
The decoder is initialized from the multimodal encoder's weights and finetuned with a language modeling loss. We restrict the decoder to generate answers from a predefined set $3k$ of candidate answers \cite{bilinear}. 

\noindent \textbf{Visual Entailment} is a visual reasoning task where a model must decide whether an image (the premise) entails a sentence (hypothesis), contradicts it, or is neutral. 
We stack a multi-layer perceptron atop the [CLS] token of the multi-modal encoder and treat the task as a 3-way classification problem.

 \noindent \textbf{Visual Grounding} requires localizing the image region corresponding to a text description (referring expression). 
 We use the RefCOCO+ dataset \cite{Yu2016ModelingCI} with $141\mathrm{k}$ referring expressions for $20\mathrm{k}$ images from the COCO dataset. 
 Following \cite{ALBEF}, we simulate a weakly supervised setting where the bounding box annotations are not used during finetuning. 
 The model is finetuned for 5 epochs in manner similar to image-text retrieval.

\begin{table}[]
    \centering
        \caption{An ablation study on the components of the proposed approach. ITM: image-text matching. ITC: image-text contrastive learning. MLM: masked language modeling. MIM: masked image modeling. PLS: pseudo-label supervision. }
    \label{tab:ablations}
\begin{tabular}{lccccccccccc}
\toprule
\multicolumn{6}{c}{Components} &  \multicolumn{6}{c}{Flickr30k True 0-shot (1k test set)} \\
\cmidrule(l{0.5em}r{0.5em}){2-6}  \cmidrule(l{0.5em}r{0.5em}){7-12}
& ITM & ITC & MLM & MIM & PLS & TR@1 & TR@5 & TR@10 & IR@1 & IR@5 & IR@10\\
\hline (a) & $\checkmark$ & & & & & 6.1 & 9.3 & 11.6 & 7.3 & 10.4 & 11.7 \\
(b) & $\checkmark$& $\checkmark$ & & & & 73.1 & 85.9 & 88.5 & 56.6 & 79.0 & 83.6\\
(c) &$\checkmark$ & $\checkmark$ & $\checkmark$ & & & 84.0 & 96.4 & 97.8 & 69.5 & 89.2 & 93.9\\
(d) & $\checkmark$& $\checkmark$& $\checkmark$ & $\checkmark$ & & 85.1 & 97.1 & 99.2 & 70.1 & 89.3 & 94.6 \\
(d) &$\checkmark$ & $\checkmark$& $\checkmark$ &$\checkmark$& $\checkmark$ & 86.2 & 97.2 & 98.7 & 69.5 & 90.2 & 94.7 \\
\bottomrule
\end{tabular}
\end{table}
\subsection{Results and Discussion}
We run each fine-tuning experiment five times with different random seeds, and report the mean and standard deviation in the following tables.

\subsubsection{Zero-shot Retrieval}
Table \ref{tab:zshot-flickr} reports results on zero-shot image-text retrieval.
Our SIMLA model outperforms both CLIP \cite{clip} and ALIGN \cite{align}, which were trained on 100x and 300x more pairs respectively.
We achieve better Rank-1 performance on both text and image retrieval compared to ALBEF \cite{ALBEF}.

\begin{table}[]
    \centering
        \caption{Zero-shot image-text retrieval results on Flickr30K.}
    \label{tab:zshot-flickr}
\begin{tabular}{lc|cccccc}
\hline \multirow{2}{*}{ Method } & \# Pre-train & \multicolumn{6}{|c}{ Flickr30K (1K test set) } \\
& Images & & TR & & & IR & \\
\hline & & R@1 & R@5 & R@10 & R@1 & R@5 & R@10 \\
UNITER  & $4 \mathrm{M}$ & $83.6$ & $95.7$ & $97.7$ & $68.7$ & $89.2$ & $93.9$ \\
CLIP  & $400 \mathrm{M}$ & $88.0$ & $98.7$ & $99.4$ & $68.7$ & $90.6$ & $95.2$ \\
ALIGN  & $1.2 \mathrm{~B}$ & $88.6$ & $98.7$ & $99.7$ & $75.7$ & $93.8$ & $96.8$ \\
ALBEF & $4 \mathrm{M}$ & $90.5$ & \textbf{98.8} & \textbf{99.7} & $76.8$ & $93.7$ & \textbf{96.7} \\
\midrule
SIMLA & $4 \mathrm{M}$ & \textbf{91.9} & 98.6 & 99.1 & \textbf{78.1} & \textbf{93.9} & \textbf{96.7} \\
Std. Dev & & $\pm0.4$ & $\pm0.4$ & $\pm0.2$ & $\pm0.4$ & $\pm 0.3$ & $\pm0.2$ \\
\hline
\end{tabular}
\end{table}
\subsection{Image-Text Retrieval}
Table \ref{tab:retrieval} reports results on fine-tuned image-text retrieval. 
Our SIMLA model outperforms all other approaches on Rank-1 retrieval across both modalities and dataset, with a substantial ($3\%$) increase over ALBEF \cite{ALBEF} and a ($6\%$) increase over OSCAR \cite{oscar} on Rank-1 MSCOCO text retrieval.
\begin{table}[t]
    \centering
        \caption{Fine-tuned image-text retrieval results on Flickr30K and MSCOCO.}
    \label{tab:retrieval}
    \begin{tabular}{lc|cccccc|cccccc}
\hline
\multirow{2}{*}{Method} & \multirow{2}{*}{Pairs} &  \multicolumn{6}{c|}{Flickr30K (1k test set)} & \multicolumn{6}{c}{MSCOCO (5k test set)} \\
& & & \multicolumn{1}{c}{ TR } & &  \multicolumn{3}{c|}{ IR } & \multicolumn{3}{c}{ TR } & \multicolumn{3}{c}{ IR } \\
\hline & & $\mathrm{R} @ 1$ & $\mathrm{R} @ 5$ & $\mathrm{R} @ 10$ & $\mathrm{R} @ 1$ & $\mathrm{R} @ 5$ & $\mathrm{R} @ 10$ & $\mathrm{R} @ 1$ & $\mathrm{R} @ 5$ & $\mathrm{R} @ 10$ & $\mathrm{R} @ 1$ & $\mathrm{R} @ 5$ & $\mathrm{R} @ 10$ \\
UNITER & $4 \mathrm{M}$ & $87.3$ & $98.0$ & $99.2$ & $75.6$ & $94.1$ & $96.8$ & $65.7$ & $88.6$ & $93.8$ & $52.9$ & $79.9$ & $88.0$ \\
VILLA & $4 \mathrm{M}$ & $87.9$ & $97.5$ & $98.8$ & $76.3$ & $94.2$ & $96.8$ & $-$ & $-$ & $-$ & $-$ & $-$ & $-$ \\
OSCAR & $4 \mathrm{M}$ & $-$ & $-$ & $-$ & $-$ & $-$ & $-$ & $70.0$ & $91.1$ & $95.5$ & $54.0$ & $80.8$ & $88.5$ \\
ALBEF & $4 \mathrm{M}$ & $94.3$ & $99.4$ & \textbf{99.8} & $82.8$ & \textbf{96.7} & \textbf{98.4} & $73.1$ & $91.4$ & $96.0$ & $56.8$ & $81.5$ & \textbf{89.2} \\
\hline
SIMLA & $4 \mathrm{M}$ & \textbf{94.7} & \textbf{99.5}  & 99.7 & \textbf{83.3}  & 96.5 & 98.2 & \textbf{75.8} & \textbf{92.9} & \textbf{96.2} & \textbf{57.7} & \textbf{81.9}  & \textbf{92.0}  \\
Std. Dev &  & $\pm0.2$ & $\pm0.1$  & $\pm0.1$ & $\pm0.2$ & $\pm0.2$ & $\pm0.1$ & $\pm0.3$ & $\pm0.3$ & $\pm0.1$ & $\pm0.2$ & $\pm0.4$  & $\pm0.55$  \\
\hline
\end{tabular}
\end{table}
\subsection{VQA/NLVR/SNLI-VE}
Table \ref{tab:results} compares the performance of SIMLA to existing methods on vision-language understanding tasks.
SIMLA achieves state of the art performance, outperforming methods that use object annotations \cite{oscar} adversarial training \cite{villa}, and dual stream architectures \cite{ALBEF}.
\begin{table}[!t]
	\caption
	{
	\small	
		Comparison with state-of-the-art methods on downstream vision-language tasks.
	ALBEF results are from our reproduction, due to expired URLs in NLVR. 
 SNLI-VE results may be noisy due to label errors \cite{corrected_snli_ve}.
	}
	\label{tab:results}
    \aboverulesep = 0.48mm
    \belowrulesep = 0.48mm
    \small
	\centering	
	\begin{tabular}	{l  l l  l  l  l    l  }
		\toprule	 	
	 \multirow{2}{*}{Model} & \multicolumn{2}{c}{VQA} & \multicolumn{2}{c}{NLVR$^2$} & \multicolumn{2}{c}{SNLI-VE} \\
        \cmidrule(l{0.5em}r{0.5em}){2-3} \cmidrule(l{0.5em}r{0.5em}){4-5} 
        \cmidrule(l{0.5em}r{0.5em}){6-7} 
	  & test-dev & test-std & dev & test-P & val & test\\
	  \midrule
	  VisualBERT~\cite{VisualBERT} & 70.8 & 71.0 & 67.4 & 67.0 & - & - \\
	  VL-BERT~\cite{VL-BERT} & 71.16 & - &  - & - & - & - \\
	  LXMERT~\cite{LXMERT}  & 72.4 & 72.5 & 74.9 & 74.5 & - & - \\
	  12-in-1~\cite{12in1} & 73.2 & - & - & 78.9 & - & 77.0 \\
	  UNITER~\cite{uniter} & 72.7 & 72.9 & 77.2 & 77.9 & 78.6 & 78.3 \\
	   VL-BART/T5~\cite{VL_T5} & - & 71.3 & - & 73.6& - & - \\
	   ViLT~\cite{ViLT}  & 70.9 & - & 75.2 & 76.2 & - & - \\
	   OSCAR~\cite{oscar} &  73.2 & 73.4 & 78.1 & 78.4 & - & - \\
	   VILLA~\cite{villa} & 73.6 & 73.7 & 78.4 & 79.3 & 79.5 & 79.0 \\
	  ALBEF~\cite{ALBEF} & 74.5 & 74.7 & 79.2 & \textbf{80.0} & 79.1 & 80.1 \\
	  \midrule
	   SIMLA & \textbf{74.5} & \textbf{74.8} & \textbf{79.8} &  79.5 & \textbf{79.6} &  \textbf{80.2} \\
          Std. Dev & $\pm0.1$ & $\pm0.1$ & $\pm0.4$ &  $\pm0.5$ & $\pm{0.2}$ &  $\pm{0.3}$ \\
		\bottomrule
	\end{tabular}
\end{table}	
\subsection{Weakly-Supervised Visual Grounding} 
\begin{figure}[t]
    \centering
    \includegraphics[width=\textwidth]{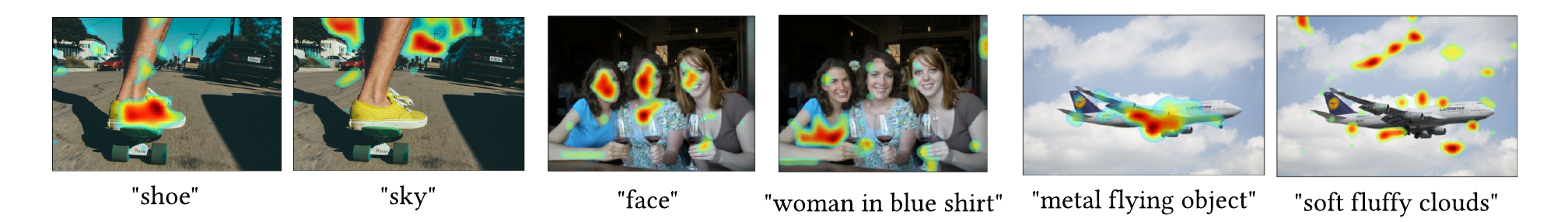}
    \caption{Examples of fine-grained alignment learned by SIMLA. The model can ground abstract concepts (e.g. metal flying object) in addition to simple concepts (e.g shoe). }
    \label{fig:attn}
\end{figure}

We show results on RefCOCO+ in Table \ref{tbl:grounding}.
We outperform ALBEF \cite{ALBEF}, which itself outperforms existing methods by $\approx 10\%-30\%$, by $1.5\%$ and $1.2\%$ on TestA and TestB respectively.
We ground a referring expression in an image using Grad-CAM \cite{grad_cam} on the cross-attention maps in the 8th layer of the multimodal encoder, using the gradients of the image-text matching score $\boldsymbol{p}^{\mathrm{itm}}(I, T)$ for a text-image pair $(I, T)$.
In Figure \ref{fig:attn}, we visualize the Grad-CAM to show the grounding and fine-grained alignment ability of the model.
\subsubsection{Data Efficiency}
\begin{figure}[th]
    \centering
    \includegraphics[width=\textwidth]{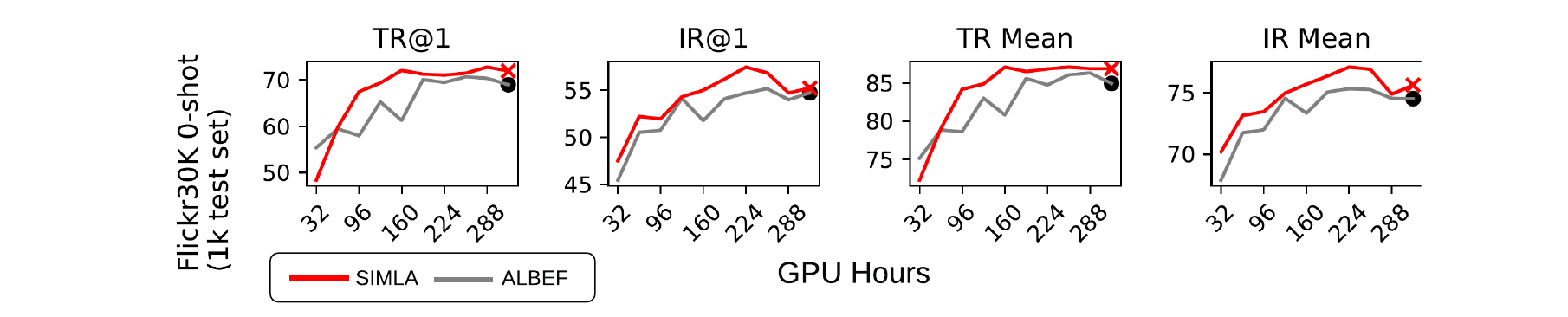}
    \caption{True zero-shot fast retrieval accuracy on the Flickr30K test set as a function of training time. SIMLA achieves higher accuracy in less training time than ALBEF.}
    \label{fig:data-efficiency}
\end{figure}

The additional pretraining tasks of SIMLA result in a stronger training signal that allow the model to learn faster with fewer training steps.
In Figure \ref{fig:data-efficiency}, we show zero-shot image-text retrieval accuracy as both ALBEF and SIMLA train.
SIMLA's zero-shot accuracy smoothly and quickly rises in the beginning stages of training, compared to the more gradual and rocky climb of ALBEF. 

\begin{table}[!t]
    \caption
	{
		\small	
		Weakly-supervised visual grounding on RefCOCO+~\cite{refer} dataset.
	}
	\label{tbl:grounding}	
	\centering	
     \setlength\tabcolsep{4pt}
		\begin{tabular}	{l  |  c c c }
			\toprule
			Method & Val & TestA & TestB \\
			\midrule
			ARN~\cite{ARN} & 32.8 & 34.4 & 32.1 \\
			CCL~\cite{CCL} & 34.3 & 36.9 & 33.6 \\
			ALBEF \cite{ALBEF} & \textbf{58.5} & 65.9 & 46.3 \\	
			\hline
			SIMLA & 58.1 & \textbf{67.4} & \textbf{47.5} \\
            Std. Dev & $\pm0.5$ & $\pm0.29$ & $\pm0.33$ \\
			\bottomrule
		\end{tabular}
\end{table}
\subsubsection{Ablation Study}
In Table \ref{tab:ablations}, we study the effect of the various losses on image-text retrieval performance.
Training with only the image-text matching (ITM) loss provides only a weak supervisory signal.
Explicit alignment is crucial, and each level of alignment provides an increase  in performance. 
Global alignment ($\mathcal{L}_{\mathrm{itc}}$) provides the largest boost in performance, but fine-grained alignment ($\mathcal{L}_{\mathrm{mim}}+\mathcal{L}_{\mathrm{mlm}}$) is crucial for increasing performance, and pseudo-label supervision ($\mathcal{L}_\mathrm{psl}$) successfully exploits additional supervisory signals to learn a better-aligned representation.
\subsubsection{Qualitative Results}
\begin{figure}
    \centering
    \includegraphics[width=\textwidth]{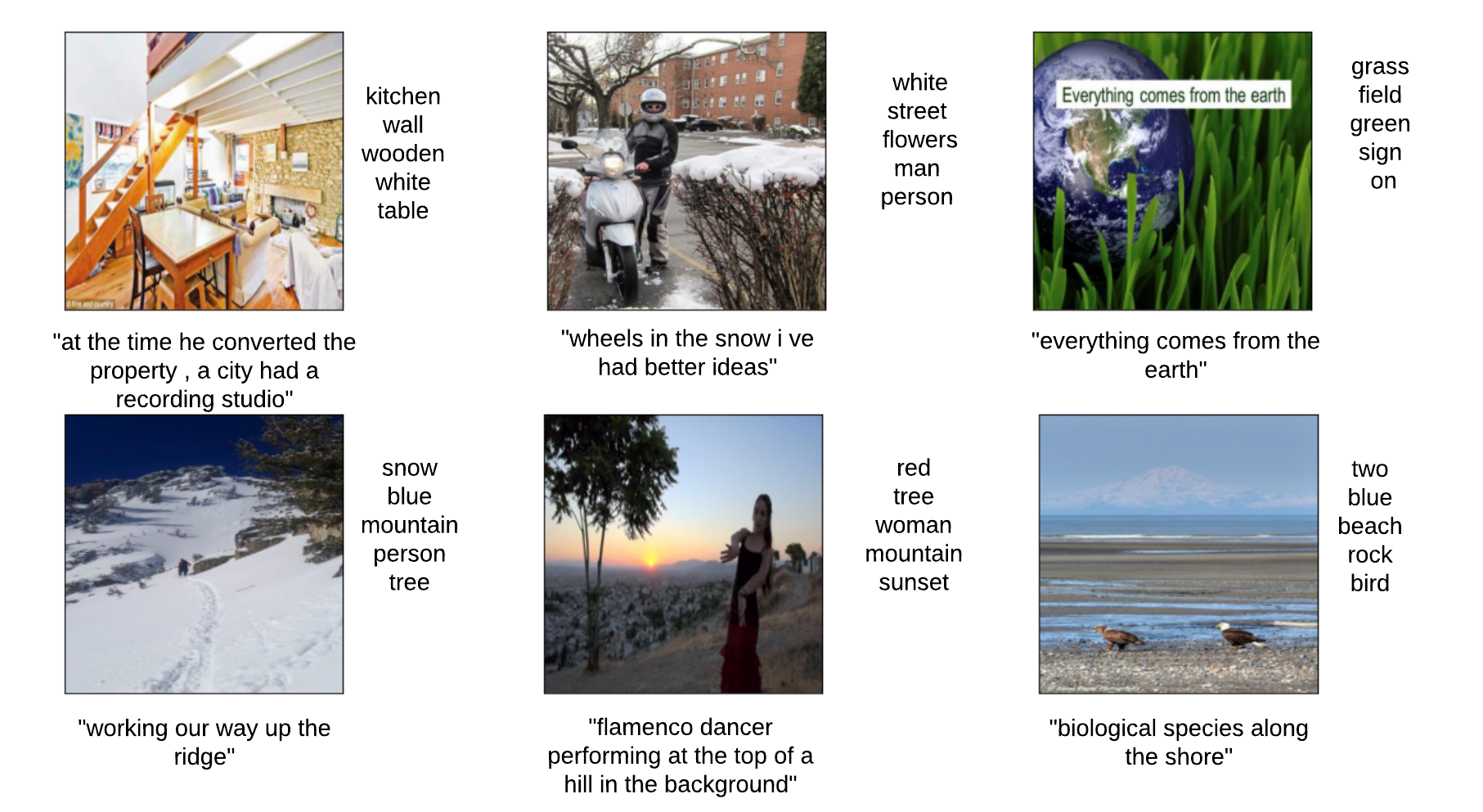}
    \caption{Examples of pseudolabels used for pseudolabel supervision, obtained by decoding high-probability concept head logits from the momentum encoders.}
    \label{fig:pseudolabels}
\end{figure}

In Figure \ref{fig:pseudolabels}, we show examples of pseudolabels generated by the momentum models. 
When the captions are nondescriptive, the pseudolabels provide a strong surrogate supervisory signal that grounds the content of the image in natural language concepts.
Even when the captions \textit{are} descriptive (bottom middle of Figure \ref{fig:pseudolabels}), the pseudolabels provide additional supervision by requiring the visual representation to reflect concepts present in the image but not in the caption.
We show self-attention maps obtained from the \texttt{[CLS]} token of the visual encoder in Figure \ref{fig:vit-attention}. 
Different heads of the visual encoder work together to decompose a scene, with heads focusing on various parts of the scene.
The subjects of attention correspond to the visual concepts humans are most likely to notice, even in cluttered or dense scenes. 
The attention map segments objects well, despite having no access to object-level annotations and receiving supervision from oftentimes noisy captions.

\subsection{Parameter Count and Inference Speed}
In Tables \ref{tab:zshot-flickr}, \ref{tab:retrieval} and \ref{tab:results}, we report results against the \texttt{large} sized versions of UNITER \cite{uniter}, OSCAR \cite{oscar} and VILLA \cite{villa}, which have $\approx335\mathrm{M}$ parameters and a depth of 24 transformer encoder layers.
SIMLA has equivalent depth (24 layers) with fewer parameters ($223\mathrm{M}$) due to parameter sharing.
SIMLA is also substantially faster at inference time (7.2 pairs / second vs $\approx 1.1$ pairs / second for UNITER/VILLA/OSCAR) due to the dual-encoder design shared with ALBEF \cite{ALBEF}, in which the text/image encoders can be used separately to quickly retrieve the top-$k$ candidates matching a query, and re-ranking them using the slower multimodal encoder. 
\section{Related Work}
\subsection{Vision Language Pretraining}
Early transformer-based \cite{transformer,bert} vision language pretraining techniques \cite{VisualBERT,ViLBERT,LXMERT,uniter} required an pretrained object detector and were limited to visual categories the object detector could identify.
Recent contrastive-image text learners such as CLIP\cite{clip} do not rely on object detectors and understand a far wider range of visual concepts.
However, CLIP relies on a massive dataset (400m pairs) to overcome label noise. 
Several methods \cite{ALBEF,declip,lit,triple_contrastive_learning,cyclip} have been proposed for data-efficient pretraining. 
DeCLIP \cite{declip} exploits inter/intra-modality supervision to train a CLIP-like model with less data, similar to \cite{triple_contrastive_learning}. 
ALBEF\cite{ALBEF} proposes a contrastive alignment followed by deeper fusion with a multimodal encoder. 
Methods such as BLIP\cite{blip}, CoCa\cite{coca}, SimVLM \cite{simvlm}, UNIMO \cite{unimo,unimo_2} incorporate a decoder and add image-to-text generation as an auxillary task.
Other lines of work on vision-language foundation models \cite{foundation_models} are multi-task models \cite{flamingo,florence,ofa,flava} or foundation model ensembles \cite{socratic_models,visual_clues,foundation_models}.
We propose a data-efficient, detector-free pretraining approach, architecturally similar to \cite{ALBEF}, but with additional supervision for the visual encoder to learn key words (high level concepts) that are present in the image, as well as a symmetric cross-modality reconstruction task inspired by the masked image modeling techniques of \cite{beit,ibot}.
\subsection{Fusion Methods}
Existing fusion techniques can be broadly classified into three categories: early \cite{avscene},\cite{slowfast}, middle \cite{epic,mmbottlenecks,mmbottlenecks}, and late fusion \cite{xdc,clip,ViLBERT,align}.
Late fusion (e.g. CLIP) is the dominant approach due to its scalability and encodes input modalities separately using unimodal encoders and fuses the resulting representations at the end.    
However, each modality can have different levels of information density.
For example, \cite{versatilenetworks} shows audio and video to have fine-grained information while text has coarse-grained information, and \cite{lit} draws the conclusion that in dual-stream contrastive architectures, the strength of the visual encoder matters more than that of the language encoder. 
It is thus essential to consider the information density of the input modalities for fusion. 
Compared to language, images require significant processing to extract useful semantic information, but current dual-stream approaches \cite{clip,lit,declip} apply the same amount of of processing to both. 
In contrast, we use a single stream architecture where the input modalities undergo asymmetric processing before fusion.
OSCAR\cite{oscar}, UNITER\cite{uniter} and VilBERT\cite{ViLBERT} also include a similar patch-level concept prediction task, but they use region labels produced by a pretrained object detector as prediction targets.
SIMLA is fully self-supervised, and needs no labels, bounding boxes, or object detectors.
\section{Conclusion}
We propose SIMLA, a framework for vision-language pretraining.
In contrast to contemporary dual-stream approaches that employ symmetric encoders and introduce multimodal interactions after unimodal representation learning, SIMLA uses a single-stream architecture with asymmetric depth of processing for each modality and enables earlier multimodal interactions. 
SIMLA aligns images and text on multiple levels, and explicitly enriches the visual modality with pseudo-label supervision to ensure similar levels of conceptual abstraction in the representations before fusion. 
We empirically verify the strength of the approach and achieve state-of-the-art results on image-text retrieval, natural language visual grounding, and vision language reasoning tasks.
Finally, we show that the additional training tasks provide additional supervision that increases the data efficiency of SIMLA relative to other state-of-the-art approaches.

\clearpage
%
%
\bibliographystyle{splncs04}
\bibliography{egbib}





\title{} 
\author{}
\institute{}
\maketitle

\pagestyle{headings}


\section{Architectural Differences}
\begin{figure}
    \centering
    \includegraphics[width=\textwidth]{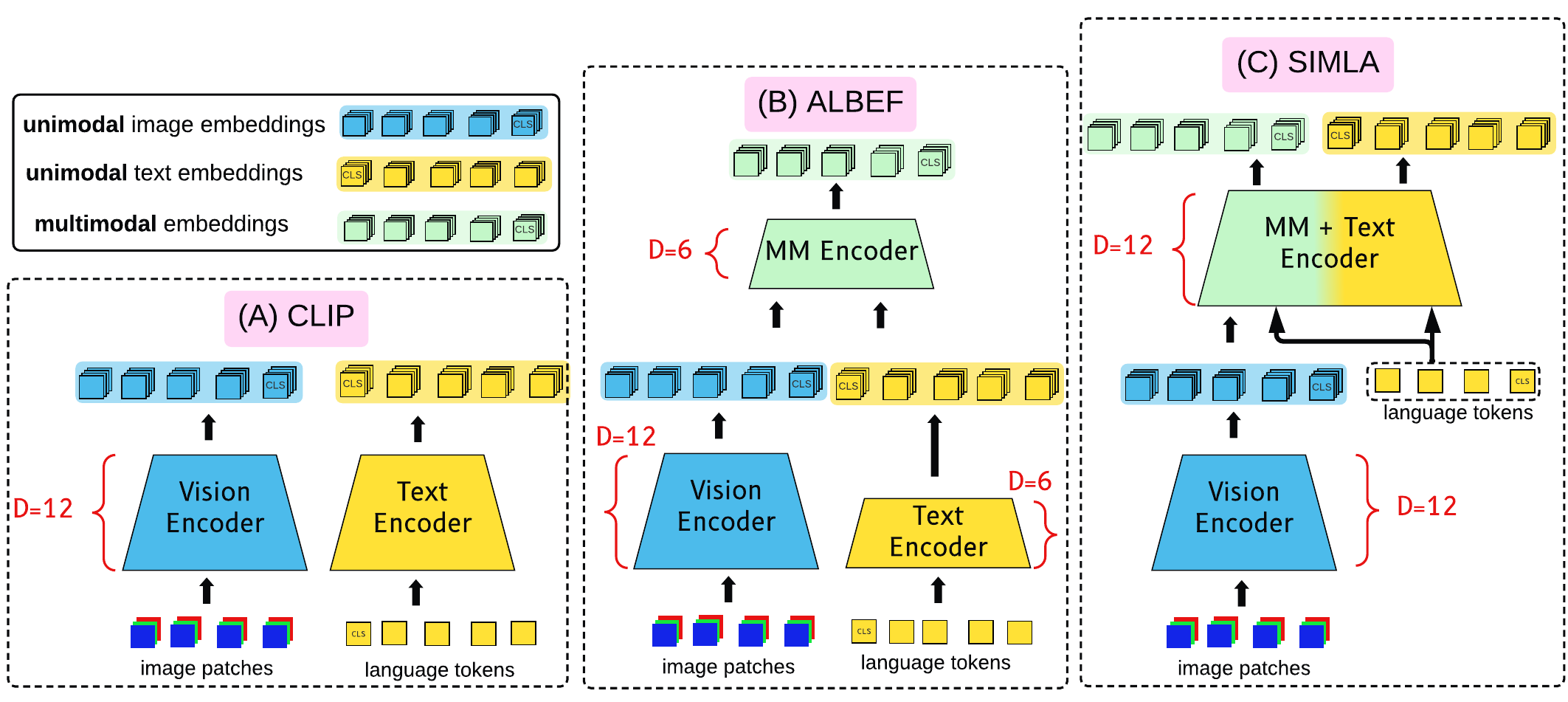}
    \caption{Architectural differences between CLIP (left panel), ALBEF (middle panel) and the proposed model (right panel).}
    \label{fig:block-diagram-arch}
\end{figure}
\subsection{CLIP vs. SIMLA}
CLIP\cite{clip} has a symmetric dual-encoder architecture which is designed for global alignment between unimodal text and image representations. 
Each encoder is a 12-layer transformer encoder dedicated to a single modality.
SIMLA has a single-stream architecture designed for alignment at multiple levels.
The primary architectural differences between CLIP and SIMLA are:
\begin{enumerate}
    \item SIMLA includes a multimodal encoder with cross-attention that enables alignment between patch-level image regions and the caption. 
    \item SIMLA adds additional training tasks, taking advantage of the multimodal encoder to align on multiple levels.
    \item SIMLA's text encoder is dual-purpose: it is used as both a multimodal encoder and text encoder by sharing weights.
\end{enumerate}
\subsection{ALBEF vs. SIMLA}
ALBEF\cite{ALBEF} can be seen as an asymmetric variant of CLIP, with a transformer-based multimodal encoder atop the unimodal text and image encoders for stronger fusion.
Furthermore, ALBEF aligns the unimodal text and image representations before fusion within the multimodal encoder.
The primary architectural differences between ALBEF \cite{ALBEF} and SIMLA are:
\begin{enumerate}
    \item SIMLA's multimodal encoder can fuse raw, unaligned language tokens with image patch embeddings from the visual encoder.
    In contrast, ALBEF's multimodal encoder requires already aligned vision/language \textit{features} as input for fusion.
    \item SIMLA reuses the multimodal encoder as a text encoder by sharing weights. 
    \item SIMLA's multimodal encoder is capable of using both image patches and language tokens as queries in the attention layers due to the cross-modality reconstruction task. 
    ALBEF's multimodal encoder can only use language tokens as queries in the attention layers.
    \item SIMLA has twice the depth of multimodal fusion (12 layers vs 6 layers) with the same number of parameters.
\end{enumerate}
\subsection{General Similarities and Differences}
ALBEF, CLIP, and SIMLA all have the same number of transformer encoder layers (24), though they are distributed differently.
Specifically, all of CLIP's layers are dedicated to unimodal representation learning (image / text encoders), with no fusion layers.
ALBEF incorporates a multimodal encoder (fusion layers), but reduces the amount of layers dedicated to unimodal representation learning (text / image encoders).
SIMLA incorporates a multimodal encoder (fusion layers), but avoids the need to reduce the number of layers dedicated to unimodal representation learning through weight sharing between the text encoder and multimodal encoder.
We take advantage of the observation \cite{ViLBERT,universalcomputation,frozenlanguagemodels} that pretrained language models have substantial capability for reuse and novel tasks, and reuse the text encoder as a multimodal encoder by adding cross-attention layers to the language model. 

\section{More Fine-Grained Alignment Examples}
\begin{figure}
    \centering
    \includegraphics[width=\textwidth]{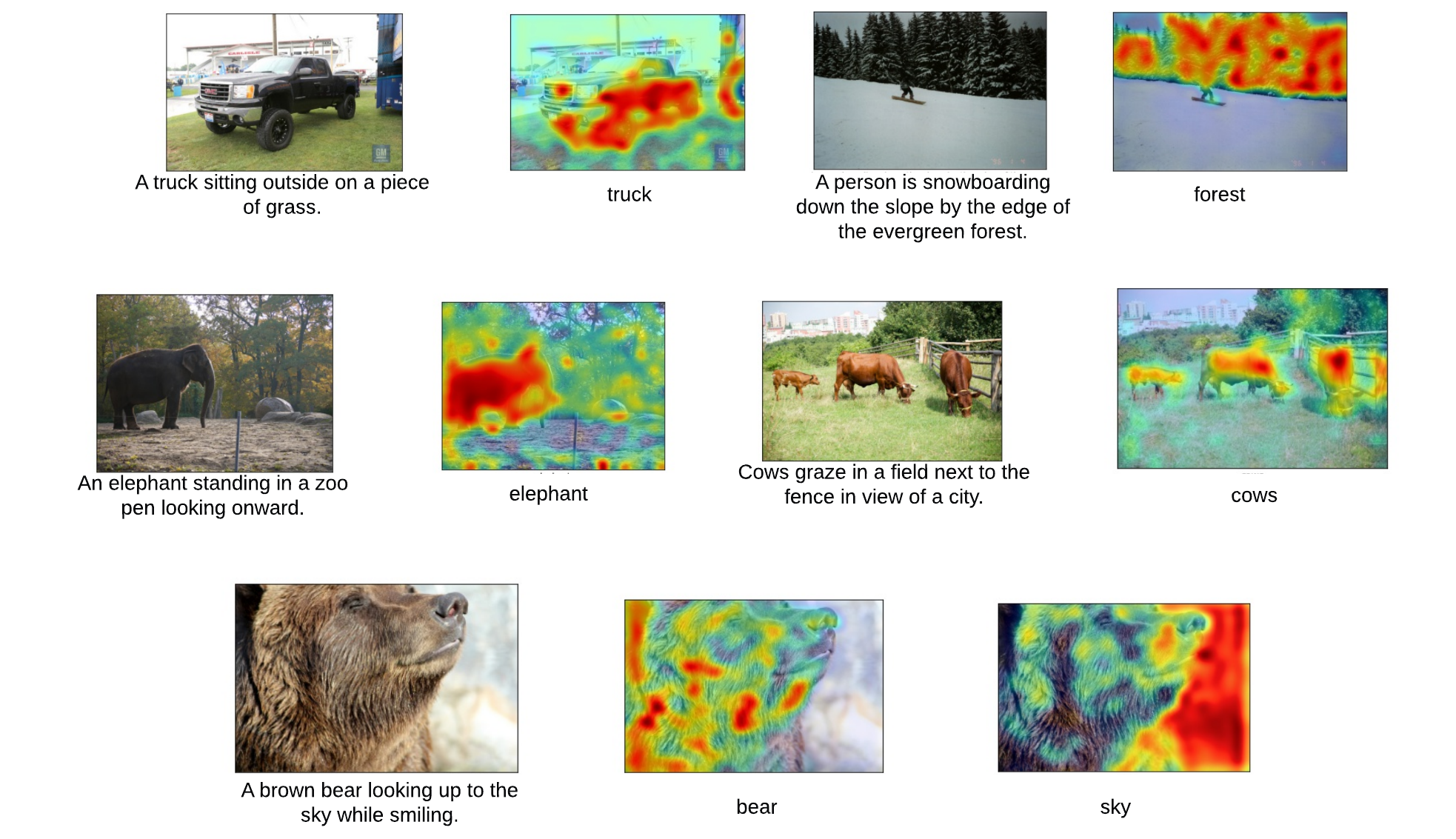}
    \caption{Grad-CAM of the image encoder through the concept prediction head.}
    \label{fig:vit-caption-grad-cam}
\end{figure}

In Figure \ref{fig:vit-caption-grad-cam}, we present examples of the image encoder's ability to ground image regions to language.
The concept head atop the image encoder's $\texttt{[CLS]}$ token is a linear classifier that predicts the presence or absence of tokens in the caption, based only on the image content.
We apply Grad-CAM \cite{gradcam} to show what image regions the image encoder is looking at when it predicts the presence of a token.
As visible in the Grad-CAM visualizations of Figure \ref{fig:vit-caption-grad-cam}, the image encoder itself is capable of rudimentary natural language grounding.
\section{Pseudolabel Extraction}
The pseudolabel supervision loss is designed to train the image encoder's representation to explicitly encode the presence of crossmodal concepts.
While "concept" is a broad term, we use it in a narrower sense: to denote object-level semantic regions of images or text.
A subset of language tokens can clearly be used to denote objects (e.g. 'cow', 'chair', 'horse').
Other language tokens have no obvious visual counterpart (e.g. 'forever', 'famine').
Based on this intuition, we use the language tokens present in a caption as labels for the associated image.
However, only a subset of these labels will correspond to cross-modality concepts which can be represented both visually and textually.
To select the subset of language tokens in a caption corresponding to cross-modal concepts, we use the attention weights of the last layer of the multimodal encoder. 
Let $\{\texttt{[cls]}, t_1,...,t_N\}$ be the input language tokens, and let $(\{\Vec{v}_\mathrm{cls}, \Vec{v}_1,...,\Vec{v}_N\})$ be the sequence of image patch embeddings produced by the image encoder for an image-text pair.
We perform a forward pass through the multimodal encoder $E_{mm}$ using the language tokens as the queries\footnote{Using image patches as queries resulted in lower quality pseudolabels.} and the image patches as the keys and values.
Using the standard formulation of cross-attention \cite{ViLBERT,transformer} in Equation \eqref{eq:cross-attention},
\begin{equation}
\label{eq:cross-attention}
\operatorname{Cross-Attention}(Q_t, K_i, V_i)=\operatorname{softmax}\left(\frac{Q_t K^{T}_i}{\sqrt{d_{k}}}\right) V_i
\end{equation}
where $Q_t$ is query embedding sequence of the language tokens, $V_i$ is the value embedding sequence of the image patches, and $K_i$ is the key embedding sequence of the image patches, we compute a series of multimodal embeddings $\{\Vec{m}_\mathrm{cls}, \Vec{m}_1,...,\Vec{m}_N\}$ having the same length as the sequence of language input tokens $\{\texttt{[cls]}, t_1,...,t_N\}$.
Next, we apply \textit{self-attention}
\begin{equation}
\label{eq:self-attention}
\operatorname{Self-Attention}(Q_{mm}, K_{mm}, V_{mm})=\operatorname{softmax}\left(\frac{Q_{mm}K^{T}_{mm}}{\sqrt{d_{k}}}\right) V_{mm}
\end{equation}
on the sequence of multimodal embeddings $\{\Vec{m}_\mathrm{cls}, \Vec{m}_1,...,\Vec{m}_N\}$, which produces an attention matrix $\mathcal{A}_{self}$ of dimensions $N \times N$, where $N$ is the length of the language sequence.
It is then straightforward to choose the top $k$ most attended positions using the 0-th row of $\mathcal{A}_{self}$, which corresponds to $\vec{m}_{\mathrm{cls}}$, the multimodal representation of the image-text pair.
The tokens in the most attended positions are then taken to be the natural language concepts most relevant to the content of the image, and are used as pseudolabels.
In practice, we found that $k=4$ yielded the best results.

\section{How much does unimodal pretraining matter?}
We experiment with training from scratch instead of initializing from pretrained weights of BERT and DeiT in Table \ref{tab:scratch-vs-pretrain}.
Initializing from pretrained core models is efficient: training from scratch slows down pretraining.
This effect will likely diminish as the number of training pairs increases.

\begin{table}[h]
	\caption
	{
	Training with pretrained core models is more efficient.
	}
	\label{tab:scratch-vs-pretrain}

    \small
	\centering	

	\begin{tabular}	{lc|llll}
		\toprule	 	

      & &  \multicolumn{2}{c}{Flickr 0-shot} & \multicolumn{2}{c}{RefCOCO+} \\
      Weight initialization  & Pairs  & TR@1 & IR@1 & TestA & TestB \\
	  \midrule
      From pretrained BERT/DeiT & 591k & 61.0 & 45.9 & 36.8 & 30.4 \\
      From Scratch & 591k  & 18.8 & 13.9 & 18.4 & 14.4 \\

		\bottomrule
	\end{tabular}
 
\end{table}	
\section{Fashion Image Retrieval}
We compare a fine-tuned version of SIMLA against the state of the art KaleidoBERT \cite{kaleido_bert} on image-text retrieval in the fashion domain using the FashionGen \cite{fashiongen} dataset.
We use original test split and follow FashionBert's \cite{fashionbert} procedure to create the gallery for evaluation.
Specifically, we sample $1000$ product IDs, and use the frontal pose for each product as the image.
For the text, we use both the product name and the product description.
We use the same fine-tuning settings as for Flickr.

\begin{table}[h]
    \centering
        \caption{Image-text retrieval on FashionGen\cite{fashiongen}.}
    \label{tab:fashion}
\begin{tabular}{l|ccc|ccc}
\toprule
Model &  TR@1 & TR@5 & TR@10 & IR@1 & IR@5 & IR@10\\
\midrule
KaleidoBERT \cite{kaleido_bert} &  33.8 & 60.6 & 68.6 & 28.0 & 60.1 & 68.4 \\
SIMLA & \textbf{48.9} & \textbf{80.2} & \textbf{89.6} & \textbf{51.3} & \textbf{82.6} & \textbf{89.9} \\
\midrule
 $\Delta$~Change & $\uparrow$~14.9 & $\uparrow$~19.6 & $\uparrow$~21.3 & $\uparrow$~23.1 & $\uparrow$~22.5 & $\uparrow$~21.5 \\
\bottomrule
\end{tabular}
\end{table}




%
%

\end{document}